\documentclass[11pt]{article}

\usepackage{emnlp2022}

\usepackage{newtxmath}
\usepackage{latexsym}
\usepackage{hyperref}
\usepackage{float}
\usepackage{lscape}

\usepackage{graphicx}
\usepackage{amsfonts,amsmath}

\usepackage{microtype}
\title{A New Pair of GloVes}
\author{Riley Carlson \andcomma John Bauer \and Christopher D. Manning \\
  Stanford NLP Group \\
  Stanford University \\
  353 Jane Stanford Way, Stanford CA 94305-9035, U.S.A. \\
  \texttt{\{rileydc, horatio, manning\}@stanford.edu}}
\begin{document}
\maketitle
\begin{abstract}
This report documents, describes, and evaluates new 2024 English GloVe (Global Vectors for Word Representation) models. While the original GloVe models built in 2014 have been widely used and found useful, languages and the world continue to evolve and we thought that current usage could benefit from updated models. Moreover, the 2014 models were not carefully documented as to the exact data versions and preprocessing that were used, and we rectify this by documenting these new models. We trained two sets of word embeddings using Wikipedia, Gigaword, and a subset of Dolma. Evaluation through vocabulary comparison, direct testing, and NER tasks shows that the 2024 vectors incorporate new culturally and linguistically relevant words, perform comparably on structural tasks like analogy and similarity, and demonstrate improved performance on recent, temporally dependent NER datasets such as non-Western newswire data.  
\end{abstract}
 
\section{Introduction}

Neural semantic word vector space models represent each word with a real-valued vector called a word embedding. One widely used set of word embeddings are the GloVe word embeddings introduced by \citet{pennington-etal-2014-glove}. This algorithm leverages a global cooccurrence matrix built with a focus on local context. The GloVe model trains on this global matrix with the resulting weights being the word embeddings in which similar words are grouped closer together in the vector space. The resulting word embeddings encode semantic relationships in a dense vector space, making them useful for various NLP tasks. Despite the rise of transformer-based models, pre-trained static embeddings like GloVe remain valuable for low-resource settings, computationally efficient models, and interpretability-focused applications.

Our motivations for updating the word embeddings with more recent data are that new words have emerged and existing words have shifted semantic meaning since the original training in 2014. For example, `covid' does not have a representation in the 2014 embeddings. Embeddings with this updated lexicon have many benefits when used in downstream tasks such as reducing out-of-vocabulary issues. In order to reflect the current usage of English words, new embeddings trained on recent language are needed. 

In this work, we incorporate a Minimum Frequency Threshold (MFT) into the vocabulary selection process for training updated word embeddings, building on insights from the work on GloVe-V \citep{vallebueno-etal-2024-statistical}. The use of an MFT allows us to strike a balance between filtering out excessively rare and noisy words while retaining less frequent but contextually important terms. The GloVe-V framework extends this approach by introducing statistical uncertainty estimates, which account for variability in embedding positions due to data sparsity. This enables training word vectors that are not only robust and expressive but also better suited for downstream tasks where rare words often hold critical importance, ensuring adaptability to modern language usage \cite{vallebueno-etal-2024-statistical}.

Through this paper, we detail the exact training procedures and demonstrate that the 2024 word embeddings have an updated lexicon reflecting today's language usage and cultural trends. They perform comparably to the 2014 embeddings on word analogy and similarity tasks, indicating similar structural and core semantic expressiveness. Furthermore, the 2024 embeddings show improved performance on temporally dependent Named Entity Recognition (NER) datasets.

In this report, we first describe the training data used to create two different sets of word embeddings, including the chosen corpora and preprocessing steps. Then, we outline the training process for easy reproducibility. Finally, we present evaluation metrics for the embeddings, including new vocabulary coverage, direct evaluation, and performance in downstream tasks.

\section{Data: 2014 vs.\ 2024}

\begin{table}[h!t]
\centering
\begin{tabular}{p{0.45\columnwidth}|p{0.45\columnwidth}}
\multicolumn{1}{c}{2014 Data} & 
\multicolumn{1}{c}{2024 Data}        \\ \hline                                       
Wikipedia \& Gigaword -- 6 billion tokens)                        & Wikipedia (2024) \& Gigaword (5th edition) -- 11.9 billion                                     \\
Common Crawl -- \newline 42 and 840 billion                & Dolma subset -- \newline 220 billion                                                                    \\ 
Twitter -- 27 billion                                     &                                                                   \\
\end{tabular}
\caption{
Comparison of the data sources used to train word embeddings in 2014 and 2024, including corpus sizes (in billions of tokens).
}
\label{data}
\end{table}

For the 2024 embeddings, we use 3 corpora to train 2 sets of embeddings: Wikipedia, Gigaword, and Dolma. For ample comparison to the 2014 Wikipedia and Gigaword embeddings, we adopt the same corpora with an updated Wikipedia dump. The Wikipedia corpus, a dataset of Wikipedia articles, is a useful source for word definitions in a more naturally occurring environment than, say, a dictionary. Along with Wikipedia, Gigaword \citep{gigaword} was used. Specifically for the 2024 vectors, we used the 5th edition of Gigaword. This corpus consists of English newswire from 4 to 7 distinct international news outlets (depending on the year) between 1994--2010. Since 2014, the Wikipedia dumps have roughly doubled in the number of tokens. To rebalance this growth, we put two copies of Gigaword in the training corpus. 

\begin{table}[h!t]
\centering
\begin{tabular}{c|c|c}
\multicolumn{1}{c}{Dataset} & \multicolumn{1}{c}{\shortstack{Percent \\ Taken}} & \multicolumn{1}{c}{\shortstack{Tokens \\ (Billions)}}\\
\hline
\begin{tabular}[c]{@{}c@{}}Common\\ Crawl\end{tabular}      & 5\%                                                      & 87.2                                                         \\
C4                                                          & 40\%                                                     & 60                                                           \\
Reddit                                                      & 100\%                                                    & 68.9                                                         \\
\begin{tabular}[c]{@{}c@{}}Project\\ Gutenberg\end{tabular} & 100\%                                                    & 2.3                                                         \end{tabular}
\caption{Dolma Training Subset}
\label{table:dolma-subset-dataset}
\end{table}

In addition to the aforementioned corpora, we also utilize Dolma v1.6 \cite{dolma}. This corpus, which was released in January of 2024, consists of 3 trillion tokens from books, programming scripts, reference materials, scholarly articles, and online content. We take a subset of over 1TB from Dolma. Table~\ref{table:dolma-subset-dataset} shows the subsets of Dolma taken and the number of tokens used. Specifically, we have web pages from Common Crawl and C4 \cite{c4}, books from Project Gutenberg, and social media from Reddit. C4 consists of data up to 2019 while the others are up to 2023. 

\section{Methods}
First, we will describe the three corpora used to train the different embeddings in more detail, in addition to any preprocessing steps taken. Next, we will describe the training process used for all embeddings. Then, the different evaluation experiments will be presented. 

\subsection{Corpus \#1: Wikipedia and Gigaword}
 The Wikipedia part of the training corpus 
 for the 2024 vectors was downloaded from the Wikipedia dump at \url{https://dumps.wikimedia.org/enwiki/20240720/enwiki-20240720-pages-meta-current.xml.bz2}. The data was then extracted using Wiki\-Ex\-tractor \citep{Wikiextractor2015}. The Wikipedia data was cleaned by removing tags such as \texttt{<doc>} and \texttt{<unk>} tokens. 

The data was preprocessed using Stanford's CoreNLP tokenizer (version 4.4.1)\footnote{\url{https://nlp.stanford.edu/software/tokenizer.html}} using lowercase letters. The Wikipedia and Gigaword corpora were then merged with Gigaword being included twice. Together, this corpus is about 60GB with Gigaword accounting for about 74\% and Wikipedia accounting for the rest.

The vocabulary size for the Wiki/Giga vectors was selected following the methodology outlined by \citet{vallebueno-etal-2024-statistical}. Specifically, the size of the vocabulary was determined by setting a MFT for words to be included in the corpus. Through experiments with vectors trained using different MFTs, it was observed that an MFT of 20 yielded the highest average cosine similarity between the trained vectors and their Weighted Least Squares (WLS) vectors. A high cosine similarity with the WLS vectors indicates that the trained embeddings closely align with the statistically optimal solution derived from the cooccurrence matrix, reflecting robust and accurate word representations \citep{vallebueno-etal-2024-statistical}. For this corpus, using an MFT of 20 resulted in a vocabulary size of 1,291,146 words.

\subsection{Corpus \#2: Dolma}
Like the other corpora, we preprocessed using Stanford's CoreNLP tokenizer in the same manner. After preprocessing, we removed \texttt{<unk>} tokens. A maximum vocabulary size of 1.2 million was used. The vocabulary building process was done independently on the different subsets of Dolma and merged at the end to have 1.2 million vocabulary size. Further, the cooccurrence matrix was created by merging the cooccurrence matrices on the merged vocabulary. 

\subsection{Training}
The training process was consistent across all embeddings and corpora. For each embedding, the vocabulary and cooccurrence matrix were first constructed. The vocabulary size was set to over 1.2 million for the Wikipedia and Gigaword corpus and 1.2 million for the Dolma corpus. A symmetric context window of size 10 was used to define cooccurrences. Once the cooccurrence matrix was built, it was shuffled with a fixed seed of 123 for the Wiki/Giga matrix and 2024 for Dolma matrix. 

Embeddings of dimensions 50, 100, 200, and 300 were trained for the Wikipedia and Gigaword corpus, and 300-dimensional embeddings were trained for the Dolma corpus. The embeddings were optimized using GloVe’s original optimizer, AdaGrad. The training process was executed using the \texttt{demo.sh} script provided in the \href{https://github.com/stanfordnlp/GloVe}{GloVe repository} with more documentation in the \texttt{Training\_README.md} file. The hyperparameters used during training are summarized in Table~\ref{tab:hyperparams}.

\begin{table}
\centering
\begin{tabular}{l|l}
\multicolumn{1}{c}{Hyperparameter} & \multicolumn{1}{c}{Value} \\
\hline 
Learning Rate                                                                 & 0.05$^\ast$  \\
Alpha                                                                         & 0.75  \\
XMax                                                                          & 100   \\
Seed                        & 2024$^\ast$ \\

\begin{tabular}[c]{@{}l@{}}Epochs:\\ (50d, 100d)\end{tabular}         & 50    \\
\begin{tabular}[c]{@{}l@{}}Epochs: \\ (200d, 300d)\end{tabular} & 100  
\end{tabular}
\caption{Training hyperparameter summary for 50d, 100d, 200d, and 300d word embeddings trained on Wiki/Giga and 300d trained on Dolma\\
$^\ast$0.075 learning rate and 123 seed used for 50d Wiki/Giga vectors
}
\label{tab:hyperparams}
\end{table}

\subsection{Evaluation: Updated Lexicon}
To assess the quality of the embeddings, we examine the words present in the 2024 embeddings but absent from the 2014 embeddings to determine if new commonly used words are reflected in the updated embeddings. The 2014 and 2024 embedding vocabularies from the Wikipedia and Gigaword corpora were compared, as well as the 2024 Dolma embedding vocabularies with those from the 2014 840B vectors trained on Common Crawl. By representing the vocabularies as sets, we compute the difference by subtracting the 2014 set from the 2024 set. From this resulting set, we select 39 representative examples for each training corpus to illustrate our findings. 
	
\subsection{Evaluation: Direct Evaluation}
We performed direct evaluation tasks on the embeddings, comparing them with the 2014 embeddings. The evaluation focused on two primary tasks: word analogy and word similarity.

For the word analogy task, the goal is to predict a fourth word in the analogy format \texttt{word\_1 : word\_2 :: word\_3 : ?} and compare it against a gold-standard labeled word to calculate accuracy. We use two benchmark datasets:
\begin{itemize}
    \item \textbf{Google Analogy dataset} \citep{word2vec}, which comprises 8,869 semantic and 10,675 syntactic word pairs.
    \item \textbf{MSR Analogy dataset} \citep{msr}, containing 8,000 syntactic word pairs.
\end{itemize}

For the word similarity task, embeddings were evaluated by assigning similarity scores to word pairs and comparing these scores to human-annotated benchmarks. We use three benchmark datasets:
\begin{itemize}
    \item \textbf{WordSim353} \citep{ws353}, which contains 353 word pairs classified as highly similar, less similar but related, or unrelated.
    \item \textbf{SimLex999} \citep{hill-etal-2015-simlex}, comprising 999 word pairs annotated with semantic similarity scores.
    \item \textbf{MEN} \citep{MEN}, which includes 3,000 word pairs annotated with human-judged relatedness scores.
\end{itemize}

To perform these evaluations, we utilized the embedding evaluation package developed by \citet{kudkudak}, which supports analysis on both word analogy and word similarity datasets.

\subsection{Evaluation: NER}
To further assess the performance of the new embeddings, we evaluate on the downstream task Named Entity Recognition (NER).  In this task, tokens in a sentence are tagged with predefined entity categories such as persons, locations, organizations, and other proper nouns, enabling structured information extraction from text. For this evaluation, we use Stanford's Stanza NER\footnote{\url{https://stanfordnlp.github.io/stanza/ner.html}} model \citep{qi-etal-2020-stanza}, modified to replace the default word embeddings with the ones we trained. We train a model for each word embedding and dataset.  

For NER evaluation, we trained and tested models using three datasets, separately for the 2014 and 2024 embeddings:
\begin{itemize}
    \item \textbf{CoNLL-03:} Published in 2003, this dataset includes entities for persons, locations, organizations, and miscellaneous categories (entities not covered by the other categories) \citep{conll03}.
    \item \textbf{CoNLL-PP:} An improved version of CoNLL-03 introduced by \citet{liu-ritter-2023-conll}, featuring updated and modernized data. We trained models on CoNLL-03 and evaluated them on the CoNLL-PP test set.
    \item \textbf{English Worldwide Newswire:} This dataset, introduced in \citep{shan-etal-2023-english}, consists of over 1,000 English newswire articles published in 2023. These articles were sourced from 47 countries, excluding American news outlets, to ensure a non-Western focus and recent language use. The dataset includes references to major events such as the COVID-19 pandemic, providing a unique opportunity to evaluate how embeddings trained on pre-2020 data (2014 embeddings) generalize to recent contexts.
    \item \textbf{Emerging and Rare entity recognition (WNUT 17):} This 6 class dataset by \citet{wnut17} includes entities from user-generated text on platforms such as Youtube, Twitter, and Reddit. The entities are rarer and often unseen, making them challenging even for humans to tag in noisy text. This data aims to improve the detection and classification of uncommon entities in dynamic, real-world text scenarios.
\end{itemize}

The training subsets of these datasets were used to train the models with default parameters and no embedding finetuning or character language modeling, and the dev and test subsets were used for evaluation. We report F1 scores per entity and per token.

\begin{table*}[t]
\centering
\caption*{Direct Evaluation: Analogy \& Similarity}
\begin{tabular}{l|lllll}
\multicolumn{1}{c}{} & \multicolumn{2}{c}{Analogy} &
\multicolumn{3}{c}{Word Similarity} \\
\multicolumn{1}{c}{Embedding} & \multicolumn{1}{c}{Google} & \multicolumn{1}{c}{MSR} & \multicolumn{1}{c}{WordSim353} & \multicolumn{1}{c}{SimLex999} & \multicolumn{1}{c}{MEN} \\
\hline
2014 50d Wiki/Giga            & 0.462                      & 0.355                   & 0.448                          & 0.265                         & 0.652                   \\
2024 50d Wiki/Giga            & 0.455                      & 0.329                   & 0.431                          & 0.256                         & 0.637                   \\
2014 100d Wiki/Giga           & 0.631                      & 0.550                   & 0.477                          & 0.298                         & 0.681                   \\
2024 100d Wiki/Giga           & 0.601                      & 0.486                   & 0.455                          & 0.291                         & 0.672                   \\
2014 200d Wiki/Giga           & 0.698                      & 0.595                   & 0.515                          & 0.340                         & 0.710                   \\
2024 200d Wiki/Giga           & 0.696 & 0.574 & 0.480 &	0.326 & 0.688                    \\
2014 300d Wiki/Giga           & 0.717                      & 0.614                   & \textbf{0.544}                          & \textbf{0.371}                         & \textbf{0.737}                   \\
2024 300d Wiki/Giga           & \textbf{0.718}                      & 0.594                   & 0.486                          & 0.338                         & 0.690                   \\
2024 300d Dolma               & 0.708                       & \textbf{0.623}                    & 0.470                           & 0.270                          & 0.651                   
\end{tabular}
\caption{Accuracy of 2014 and 2024 embeddings on word analogy datasets Google and MSR and Spearman's Rank Correlation on word similarities datasets WordSim353, SimLex999, and MEN.}
\label{direct_eval}
\end{table*}

\section{Results }
We present results for the three evaluation metrics: updated lexicon, direct evaluation, and downstream evaluation.
\subsection{Updated Lexicon}

Here, we show qualitatively words that are present in the new 2024 embeddings and not in the 2014 embeddings. The new Wikipedia and Gigaword vectors provide a much larger vocabulary. Between the 2024 Wikipedia and double Gigaword and the 2014 Wikipedia and Gigaword embeddings, there were over 700K new words (not including numbers and words containing non-Latin alphabetical letters). Between the 2024 Dolma embeddings and the 2014 Common Crawl 840B embeddings, there are over 500K new words (not including numbers and words containing non-Latin alphabetical letters). In Tables~\ref{new_word_list_wiki} and~\ref{new_word_list_dolma}, we report 39 newly present words chosen by the authors that are of a cultural, political, and technological nature. 

\begin{table}[t!]
\caption*{2024 Updated Lexicon (Wikipedia)}
\begin{tabular}{lll}
afrobeats  & antiracism  & asmr           \\
binance    & bipoc      & blockchain     \\
brexit    & chatbot     & clickbait     \\
covid     & fyp & cryptocurrency \\
deepfake   & docuseries  & doja           \\
doordash   & draftkings  & rizz     \\
nonbinary    & fintech     & fortnite   \\
skibidi$^\ast$    & idk         & jungkook     \\
latinx     & lgbtqia     &  microaggression        \\
lstm       & metoo       & microplastic   \\
pickleball & retweet     & zelenskyy     \\
teladoc    & web3  & tiktok         \\
transwoman &  girlboss       & viserys        \\
\end{tabular}
\caption{39 word samples of new words included in Wiki/ Giga embeddings compared to the 2014 Wiki/Giga vectors. \\
$^\ast$ Not in Dolma vectors}
\label{new_word_list_wiki}
 \end{table}

\subsection{Word Embedding Evaluation}
We report the results of the 2014 and 2024 embeddings on word analogy and similarity datasets in Table~\ref{direct_eval}. Analogy tasks are assessed using the Google and MSR datasets, with accuracy as the metric. Word similarity tasks are evaluated on WordSim353, SimLex999, and MEN using Spearman's Rank Correlation coefficient ($\rho$).

\begin{table}[t!]
\caption*{2024 Updated Lexicon (Dolma)}
\begin{tabular}{lll}
dinkies  & profeminist$^\dag$  & theranos$^\dag$           \\
chatgpt$^\dag$    &    adagrad   & databricks$^\dag$     \\
huggingface     & tarboosh$^\dag$     &    gamestonk   \\
badbunny      & yeet$^\dag$ & patreon$^\dag$ \\
brainrot   & xgboost$^\dag$  & bytedance$^\dag$           \\
fakenews   &  periodt & duolingo$^\dag$       \\
mansplains    & pytorch$^\dag$     &    \small absofreakinglutely    \\
squidgame   & trumpism$^\dag$         &    clapback    \\
highkey     &   bffr   &      situationship     \\
cybertruck$^\dag$       & boujee$^\dag$       & alphafold$^\dag$   \\
glowup & openai$^\dag$     &   scikit  \\
bingewatch    & tensorflow$^\dag$  & kubernetes$^\dag$         \\
aapi$^\dag$ & airpods$^\dag$        &    deeplearning     \\    
\end{tabular}
\caption{39 word samples of new words included in the Dolma embeddings compared to the 2014 840B Common Crawl vectors. \\
$^\dag$ Also in Wiki/Giga vectors 
}
\label{new_word_list_dolma}
\end{table}

\begin{table*}[t!]
\small
\caption*{NER Scores: CoNLL}
\begin{tabular}{l|llll}
\multicolumn{1}{c}{Embedding} & \multicolumn{1}{c}{Per Entity (2003)} & \multicolumn{1}{c}{Per Entity (PP)} & \multicolumn{1}{c}{Per Token (2003)} & \multicolumn{1}{c}{Per Token (PP)} \\ \hline
2014 50d Wiki/Giga            & 89.52                                & 81.58                               & 89.30                                & 80.77                              \\
2024 50d Wiki/Giga            & 89.74                                 & 83.64                                & 89.43                                & 82.23                               \\
2014 100d Wiki/Giga           & 90.62                                 & 84.40                               & 90.41                                & 82.73                              \\
2024 100d Wiki/Giga           & 90.34                                 & 83.53                               & 90.16                                 & 82.04                              \\
2014 200d Wiki/Giga           & 90.88                                 & 84.21                               & 90.91                                & 82.89                              \\
2024 200d Wiki/Giga           & 90.69                                  & 84.36                                & 90.46                                 & 82.75                               \\
2014 300d Wiki/Giga           & 90.60                                 & 84.25                               & 90.43                                & 82.70                               \\
2024 300d Wiki/Giga           & \textbf{90.72}                          & 84.06                               & \textbf{90.50}                                & 82.74                              \\
2024 300d Dolma               & 90.05                                  & \textbf{85.14}                                & 90.12                                 & \textbf{83.69}                              
\end{tabular}
\caption{Average Test F1 scores per entity and per token on CoNLL-03 \cite{conll03} and CoNLL-PP \cite{liu-ritter-2023-conll} for 2014 and 2024 embeddings. We used Stanford's Stanza NER model with no embedding finetuning trained on CoNLL-03.}
\label{conll_ner}
\end{table*}

For the analogy tasks, the 2024 embeddings perform roughly similarly to the 2014 embeddings on the Google dataset, but have a slightly lower performance on the MSR dataset. Further, across both datasets, accuracy improves consistently as the dimension size increases for both 2014 and 2024 embeddings. 

For the word similarity tasks, the 2024 embeddings perform competitively with the 2014 embeddings across most datasets and dimensions. Both 2024 300-dimensional embeddings show a drop in rank correlation compared to the 2014 embeddings, particularly on SimLex999. This decline is addressed in the Discussion section.

\subsection{NER}
\begin{table}[t!]
\small
\caption*{NER Scores: Worldwide}
\begin{tabular}{l|ll}
\multicolumn{1}{c}{Embedding} & \multicolumn{1}{c}{Per Entity} & \multicolumn{1}{c}{Per Token} \\
\hline
2014 50d Wiki/Giga  & 82.1      & 81.04     \\
2024 50d Wiki/Giga  & 84.64      & 83.88     \\
2014 100d Wiki/Giga & 85.29      & 84.58     \\
2024 100d Wiki/Giga & 85.55      & 84.25     \\
2014 200d Wiki/Giga & 84.41       & 83.53     \\
2024 200d Wiki/Giga & 85.68      & 84.92      \\
2014 300d Wiki/Giga & 84.53      & 83.89     \\
2024 300d Wiki/Giga & 84.89      & 84.11     \\
2024 300d Dolma     & \textbf{86.23}       & \textbf{85.27}     
\end{tabular}
\caption{Average Test F1 scores per entity and per token for 2014 and 2024 embeddings on Worldwide dataset \cite{shan-etal-2023-english}. We used Stanford's Stanza NER model with no embedding finetuning trained on Worldwide dataset.}
\label{foreign_ner}
\end{table}

We evaluated the 2014 and 2024 embeddings on NER tasks using four datasets: CoNLL-03, CoNLL-PP, Worldwide, and WNUT 17. For these datasets, we report test F1 scores on a per-entity and per-token basis. The results are summarized in Tables~\ref{conll_ner}, \ref{foreign_ner}, and~\ref{wnut17}.

\begin{table}[t!]
\small
\caption*{NER Scores: WNUT17}
\begin{tabular}{l|ll}
\multicolumn{1}{c}{Embedding} & \multicolumn{1}{c}{Per Entity} & \multicolumn{1}{c}{Per Token} \\
\hline
2014 50d Wiki/Giga  & 32.95      & 31.05     \\
2024 50d Wiki/Giga  & 35.65      & 33.10     \\
2014 100d Wiki/Giga & 36.48      & 33.39     \\
2024 100d Wiki/Giga & 36.33      & 34.23     \\
2014 200d Wiki/Giga & 35.68       & 33.31     \\
2024 200d Wiki/Giga & 37.46      & 35.63      \\
2014 300d Wiki/Giga & 36.64      & 33.73     \\
2024 300d Wiki/Giga & 37.17      & 33.33     \\
2024 300d Dolma     & \textbf{39.44}       & \textbf{34.22}     
\end{tabular}
\caption{Average Test F1 scores per entity and per token for 2014 and 2024 embeddings on WNUT17 dataset \cite{wnut17}. We used Stanford's Stanza NER model with no embedding finetuning trained on WNUT17 dataset.}
\label{wnut17}
\end{table}

The results across the four NER datasets demonstrate that the 2024 embeddings generally outperform their 2014 counterparts, with particularly notable improvements on temporally dependent datasets. For the CoNLL datasets in Table~\ref{conll_ner}, the 2024 embeddings perform comparably on CoNLL-03 but show clear advantages on the modernized CoNLL-PP version, with the 2024 50d Wiki/Giga embeddings achieving the highest relative improvement in per-entity scores (83.64 vs.\ 81.58). On the Worldwide dataset in Table~\ref{foreign_ner}, the 2024 embeddings demonstrate consistent improvements across both per-entity and per-token F1 scores, with the 2024 50d Wiki/Giga embeddings achieving a per-entity F1 score of 84.64, significantly outperforming the 82.1 score of their 2014 counterparts. The challenging WNUT17 dataset in Table~\ref{wnut17} shows a significant drop in overall performance compared to CoNLL and Worldwide (F1 scores ranging between 30--40), but the 2024 embeddings consistently outperform their 2014 counterparts, with the 2024 200d Wiki/Giga embeddings achieving the highest per-entity and per-token F1 scores of 37.46 and 35.63, respectively.

Across all three datasets, the 2024 embeddings demonstrate the most pronounced gains in lower dimensions, particularly at 50d, where the differences between 2024 and 2014 embeddings are the largest. While higher dimensions (200d and 300d) achieve the best absolute F1 scores, the relative gains between 2024 and 2014 embeddings are less pronounced at these dimensions. These trends suggest that the 2024 embeddings perform as well as or better than the 2014 embeddings on datasets that align temporally with their training data, while showing marked improvements on modern, linguistically diverse datasets that better reflect contemporary language usage and cultural trends.
                  
\section{Discussion}

The 2024 word embeddings introduce a diverse range of new vocabulary that reflects cultural, technological, and linguistic shifts over the past decade, including words associated with globally significant events (`covid' and `brexit'), modern slang (`brainrot' and `periodt'), emerging technologies (`chatgpt' and `blockchain'), and popular products (`airpods'). While many slang terms are absent from the Wikipedia \& Gigaword training data because modern slang takes time to transition from social media to Wikipedia, the diverse Dolma dataset compensates by capturing informal and conversational language. The inclusion of acronyms (`idk'), linguistic blends (`absofreakinglutely'), and derivations (`retweet') reveals evolving patterns in word formation driven by digital communication and social media. By capturing these shifts, the 2024 embeddings align with current vernacular and offer practical benefits for downstream tasks, such as reducing out-of-vocabulary issues in modern datasets, while further exploration of these new words provides opportunities for linguistic and sociological research.

In the analogy tasks reported in Table~\ref{direct_eval}, the 2014 embeddings performed slightly better, with minimal differences on the Google dataset (within 0.01--0.03 accuracy) but larger gaps on the MSR dataset, particularly for lower dimensions where differences reached 0.07 for 100d and 0.03 for 50d embeddings. There were about 1100 instances that the 2014 50d predicted correctly, but the 2024 50d predicted incorrectly. About half of these errors were geography-based (e.g., cairo, egypt : bern, \textunderscore). The other half is mainly instances in which 2024 embeddings predicted a synonym of the golden answer. For example in simple, simpler: cold, \textunderscore, the 2024 embeddings predicted 'cooler' instead of 'colder'. Although the 2024 50d embeddings made these mistakes, the 2014 embeddings made these common mistakes as well. There were about 900 instances that the 2024 50d predicted correctly, but the 2014 50d predicted incorrectly. These errors were roughly half geography-based and the other using synonyms (using `knows' instead of `thinks' in write, writes : think, \textunderscore). In the errors of the geography-based ones, the 2024 embeddings do ``know'' the right answer, but the closest neighbor wasn't the right answer. For example, the 2024 embeddings do know that Bern is in Switzerland, although sometimes the closest neighbor is not Switzerland. This can be seen in 2024 embeddings not getting kabul, afghanistan : bern, \textunderscore \space right, but getting cairo, egypt : bern, \textunderscore \space correct. The same phenomenon was seen in the 2014 embeddings. The MSR dataset errors saw more instances of using a synonym in the syntactic word analogies. 

With this, the 2024 and 2014 embeddings perform roughly the same on the analogy tasks. This equal performance is expected as these tasks primarily rely on syntactic structures and commonly used words, which have remained relatively stable over the past decade.

For the word similarity tasks, there were differences in the Spearman's Rank Correlation coefficients between the embeddings. To put these difference in perspective, we looked at the word pairs for which there was a difference of $0.3$ between the embedding's prediction and the human evaluation (which was scaled to $-1$ -- 1, instead of 0--10). For the MEN dataset, there are 70 word pairs where the 2024 300d embedding's predictions fall within the threshold while the 2014 300d embeddings diverge. Conversely, there are 197 word pairs where the 2024 embeddings diverge while the 2014 embeddings remain within the threshold. For the SimLex999 dataset, these numbers are 14 and 43, respectively, and for the WS353 dataset, they are 10 and 24.

Comparing the two embedding models on MEN, the 2024 embeddings excel at capturing close synonyms and hypernym-hyponym relations (e.g., cemetery -- graveyard, stair -- staircase, ice -- snow, sea -- water). In these instances, the 2024 model aligns closely with the high similarity ratings in the dataset, while the 2014 embeddings tend to underestimate their similarity. This suggests that the 2024 embeddings may do a more precise job at clustering terms that share core, near‐equivalent meanings or clear part-whole relationships. 

\begin{table*}[t!]
\caption*{NER WNUT-17 Confusion Matrix}
\begin{tabular}{c|ccccccc}
t\textbackslash{}p & O & CORP & CREATIVE-WRK & GROUP & LOC & PER & PROD \\
\hline
O                                   & 58  & 8    & $-61$          & $-1$    & $-18$ & 8   & 6    \\
CORP                                & 0   & 5    & 0            & 2     & $-6$  & $-1$  & 0    \\
CREATIVE-WRK                        & 26  & $-3$   & $-39$          & 7     & 8   & $-2$  & 3    \\
GROUP                               & 1   & 1    & $-2$           & 4     & 1   & $-3$  & $-2$   \\
LOC                                 & $-1$  & 1    & $-7$           & $-4$    & 16  & $-4$  & $-1$   \\
PER                                 & 2   & 5    & $-7$           & 15    & $-5$  & $-10$ & 0    \\
PROD                                & 186 & 10   & 1            & 7     & 3   & 11  & 35       
\end{tabular}
\caption{2024/2014 300d Wiki/Giga confusion matrix differences on the WNUT-17 test set}
\end{table*}

\begin{table*}[t!]
    \caption*{Example NER Tagging}
    \begin{tabular}{p{0.70\textwidth}p{0.14\textwidth}p{0.14\textwidth}}
        \textbf{Sentence} & \textbf{2024} & \textbf{2014} \\
        \hline
        His repeated questioning of the system has prompted the Supreme Court to open an investigation into \textbf{Bolsonaro}. (Worldwide)   
        & PER 
        & LOC \\
        \hline
        Nationwide, \textbf{COVID-19} infections in United States are at their peak with an average of 193,863 new cases reported each day over the past week \ldots\ (CoNLL-PP)
        & MISC
        & O \\
        \hline
        Man \textbf{Finna} bring me a \includegraphics[height=0.75em]{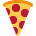} up to my job. \includegraphics[height=0.75em]{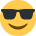} \footnotemark (WNUT-17)
        & O
        & PER \\
    \end{tabular}
    \caption{Example sentences from NER datasets with the correct tagging of the bolded word from the 2024 model is shown compared against 2014 model's tagging. }
    \label{ner_example_sentences}
\end{table*}

\begin{table}[t!]
\caption*{NER CoNLL-PP Confusion Matrix}
\begin{tabular}{c|ccccc}
 t $\backslash$ p    & O      & LOC  & MISC & ORG  & PER  \\
     \hline
O    & $-17$ & $-3$  & 5    & 18  & $-3$  \\
LOC  & $-3$  & $-19$ & 2    & 21  & $-1$  \\
MISC & 12  & $-8$  & 8    & 1   & $-12$ \\
ORG  & 13  & $-21$ & 13   & 2   & $-7$  \\
PER  & 22  & $-3$  & $-4$   & $-12$ & $-3$ 
\end{tabular}
\caption{2024/2014 300d Wiki/Giga confusion matrix differences on the CoNLL-PP test set}
\label{confusion:conllpp}
\end{table}

However, there are also cases where 2014 outperforms 2024. These often involve looser thematic or distributional associations, such as color words (blue -- red, purple -- yellow) or common everyday items that the dataset treats as reasonably similar by virtue of category or context (e.g., daffodil -- tulip, puddle -- splash, chicken -- lamb, potato -- tomato). In these scenarios, the 2024 model either overestimates or underestimates similarity, whereas 2014 captures these broader relationships more accurately.

On the WS353 dataset, we observe a similar pattern to what emerged in previous comparisons. The 2024 embeddings tend to capture near‐synonyms and clearly related category pairs more precisely. For instance, pairs such as gem -- jewel or coast -- shore have high ground‐truth similarity scores that 2024 aligns with closely, whereas the 2014 embeddings underestimate these tight connections. The same holds for other near‐synonym or hypernym-hyponym examples like magician -- wizard or lobster -- food, suggesting that 2024 more reliably encodes strong, direct lexical relationships.

Conversely, 2014 excels with looser or more contextual associations, where the concepts are functionally or thematically related rather than strictly synonymous. For example, plane -- car or energy -- laboratory receive moderately high similarity ratings in WS353, reflecting that they share an overarching category or context (transportation, scientific research). The 2024 model underestimates these relationships, implying it is not capturing certain broad or looser conceptual links as well. 

On SimLex‐999, we again see the two models excelling in different areas. The 2024 embeddings struggle with some pairs that share a domain or have moderate conceptual overlap, such as disease -- infection or river -- sea. Although these pairs are not synonyms, SimLex‐999 rates them fairly high in similarity, which 2024 underestimates more than 2014 does. Similarly, 2024 sometimes undervalues near‐synonymous verbs (e.g., deserve -- earn, remain -- retain, replace -- restore), suggesting it can lose track of subtle lexical overlaps even though those pairs have a significant degree of semantic closeness. 

Furthermore, in the top ten deviations for the 2024 Dolma 300-dimensional embeddings and the human annotations, all ten examples were the model having a high similarity between antonyms (e.g., agree -- argue). For the 2024 Wiki/Giga 300-dimensional vectors, six out of the 10 were highly similar antonyms with the other 4 being synonyms not being given a high enough similarity (e.g., creator -- maker).

Overall, across MEN, WS353, and SimLex‐999, a consistent pattern emerges: the 2024 embeddings are especially good at capturing strong, direct relationships (near-synonyms, hypernym-hyponym links, and part-whole connections) while sometimes underestimating more contextual associations. In contrast, the 2014 embeddings tend to do better on looser thematic or distributional relationships (e.g., color words, shared domains, functional overlaps) but occasionally fail on the most obvious, high‐similarity pairs like strict synonyms. 

In contrast to the direct evaluation results where it was not clear which embedding performed better, the NER evaluations reveal the 2024 embeddings outperforming the 2014 embeddings on newer and out-of-domain datasets. On the classic 2003 CoNLL dataset, the results are nearly identical, suggesting that 2014 embeddings remain sufficient for corpora that mirror their original training domain and era. However, on CoNLL-PP and the newer Worldwide and WNUT-17 datasets, the 2024 embeddings show consistent improvements.

An inspection of the confusion matrices helps illuminate why 2024 embeddings perform better on more recent data. For example, Table~\ref{confusion:conllpp} highlights how the 2024-based model reduces misclassifications of newly prominent entities into the O (no entity) category or the wrong type (e.g., labeling `COVID-19' as O). Instead, the updated embeddings capture contemporary terms and are thus more likely to assign them correct tags, often MISC, rather than leaving them unlabeled or confusing them with LOC, ORG, or PER\@. Similarly, in Table~\ref{confusion:ww}, we see that the 2024 model is less likely to confuse non-Western person names with locations, a shift that reflects more exposure to or better representation of such entities in the newer corpus.
\footnotetext{Thank you \hyperlink{https://github.com/twitter/twemoji/tree/master}{Twemoji} for the emojis!}
\begin{table}[t!]
\caption*{NER Worldwide Confusion Matrix}
\begin{tabular}{c|ccccc}
 t $\backslash$ p    & O      & LOC  & MISC & ORG  & PER  \\
     \hline
O    & $-19$                   & 24                      & 68                       & $-52$                     & $-21$                     \\
LOC  & 14                    & $-29$                     & 24                       & $-2$                      & $-7$                      \\
MISC & 6                     & $-48$                     & 138                      & $-70$                     & $-26$                     \\
ORG  & 32                    & $-36$                     & 63                       & $-30$                     & $-29$                     \\
PER  & 23                    & 1                       & $-6$                       & $-2$                      & $-16$                   
\end{tabular}
\caption{2024/2014 300d Wiki/Giga confusion matrix differences on the Worldwide test set}
\label{confusion:ww}
\end{table}

These patterns are evident in the example sentences in Table~\ref{ner_example_sentences}. In the first sentence, coming from Worldwide, the 2024 embeddings were able to tag `Bolsonaro' as a person, whereas 2014 tagged it as a location. 

The second example, coming from CoNLL-PP shows 2024 correctly labeling COVID-19, while the 2014 embeddings did not label it. This is a case in which a word that is part of recent usage should be correctly tagged. The last example from WNUT-17 shows how difficult the dataset is for tagging. Despite this, the 2024 model was able to see that `finna' is slang and did not tag it, although the 2014 model tagged it as a person. This is an example of the 2024 embeddings better capturing colloquial vernacular.

Overall, we see that the 2024 embeddings better represent current language usage, compared to the 2014 embeddings. In situations where there is temporal dependency (e.g., chatbots, NER taggers, etc.), the 2024 embeddings should be used.

\section{Conclusion }

We presented new GloVe word embeddings, introducing two sets of vectors trained on updated corpora from Wikipedia, Gigaword, and a subset of Dolma. These embeddings provide valuable new vocabulary, reflecting cultural and technological shifts over the last decade and offering linguists rich data to analyze the evolution of English, particularly in the context of social media's rising influence.

On word analogy, the new embeddings perform comparably to the 2014 embeddings, demonstrating equal structural and core semantic expressiveness. For word similarity tasks, the higher dimension 2024 embeddings occasionally overestimate the semantic similarity between antonyms. Nonetheless, the 2024 embeddings show clear advantages in temporally dependent NER datasets, such as non-Western-oriented newswire data.

These findings underline the importance of updating word embeddings to keep pace with linguistic and cultural change. The 2024 embeddings represent a meaningful advancement for modern language modeling, offering tools that are better aligned with contemporary usage. Their ability to capture recent cultural, technological, and linguistic shifts makes them particularly valuable for human-centered NLP applications, such as improving chatbot interactions and designing systems that adapt to diverse and evolving user needs.



\bibliography{anthology,custom}

\begin{thebibliography}{17}
\expandafter\ifx\csname natexlab\endcsname\relax\def\natexlab#1{#1}\fi

\bibitem[{Attardi(2015)}]{Wikiextractor2015}
Giusepppe Attardi. 2015.
\newblock {WikiExtractor}.
\newblock \url{https://github.com/attardi/wikiextractor}.

\bibitem[{Bruni et~al.(2014)Bruni, Tran, and Baroni}]{MEN}
Elia Bruni, Nam~Khanh Tran, and Marco Baroni. 2014.
\newblock Multimodal distributional semantics.
\newblock \emph{J. Artif. Int. Res.}, 49(1):1–47.

\bibitem[{Derczynski et~al.(2017)Derczynski, Nichols, van Erp, and Limsopatham}]{wnut17}
Leon Derczynski, Eric Nichols, Marieke van Erp, and Nut Limsopatham. 2017.
\newblock \href {https://doi.org/10.18653/v1/W17-4418} {Results of the {WNUT}2017 shared task on novel and emerging entity recognition}.
\newblock In \emph{Proceedings of the 3rd Workshop on Noisy User-generated Text}, pages 140--147, Copenhagen, Denmark. Association for Computational Linguistics.

\bibitem[{Finkelstein et~al.(2001)Finkelstein, Gabrilovich, Matias, Rivlin, Solan, Wolfman, and Ruppin}]{ws353}
Lev Finkelstein, Evgeniy Gabrilovich, Yossi Matias, Ehud Rivlin, Zach Solan, Gadi Wolfman, and Eytan Ruppin. 2001.
\newblock \href {https://doi.org/10.1145/503104.503110} {Placing search in context: The concept revisited}.
\newblock \emph{ACM Transactions on Information Systems - TOIS}, 20:406--414.

\bibitem[{Hill et~al.(2015)Hill, Reichart, and Korhonen}]{hill-etal-2015-simlex}
Felix Hill, Roi Reichart, and Anna Korhonen. 2015.
\newblock \href {https://doi.org/10.1162/COLI_a_00237} {{S}im{L}ex-999: Evaluating semantic models with (genuine) similarity estimation}.
\newblock \emph{Computational Linguistics}, 41(4):665--695.

\bibitem[{Jastrzebski et~al.(2017)Jastrzebski, Leśniak, and Czarnecki}]{kudkudak}
Stanisław Jastrzebski, Damian Leśniak, and Wojciech~Marian Czarnecki. 2017.
\newblock \href {https://doi.org/10.48550/ARXIV.1702.02170} {How to evaluate word embeddings? {O}n importance of data efficiency and simple supervised tasks}.
\newblock ArXiv preprint arXiv:1702.02170.

\bibitem[{Liu and Ritter(2023)}]{liu-ritter-2023-conll}
Shuheng Liu and Alan Ritter. 2023.
\newblock \href {https://doi.org/10.18653/v1/2023.acl-long.459} {Do {C}o{NLL}-2003 named entity taggers still work well in 2023?}
\newblock In \emph{Proceedings of the 61st Annual Meeting of the Association for Computational Linguistics (Volume 1: Long Papers)}, pages 8254--8271, Toronto, Canada. Association for Computational Linguistics.

\bibitem[{Mikolov et~al.(2013{\natexlab{a}})Mikolov, Chen, Corrado, and Dean}]{word2vec}
Tomas Mikolov, Kai Chen, Greg Corrado, and Jeffrey Dean. 2013{\natexlab{a}}.
\newblock \href {http://arxiv.org/abs/1301.3781} {Efficient estimation of word representations in vector space}.
\newblock ArXiv preprint arXiv:1301.3781.

\bibitem[{Mikolov et~al.(2013{\natexlab{b}})Mikolov, Yih, and Zweig}]{msr}
Tomas Mikolov, Wen-tau Yih, and Geoffrey Zweig. 2013{\natexlab{b}}.
\newblock \href {https://aclanthology.org/N13-1090} {Linguistic regularities in continuous space word representations}.
\newblock In \emph{Proceedings of the 2013 Conference of the North {A}merican Chapter of the Association for Computational Linguistics: Human Language Technologies}, pages 746--751, Atlanta, Georgia. Association for Computational Linguistics.

\bibitem[{Parker et~al.(2011)Parker, Graff, Kong, Chen, and Maeda}]{gigaword}
Robert Parker, David Graff, Junbo Kong, Ke~Chen, and Kazuaki Maeda. 2011.
\newblock \href {https://doi.org/https://doi.org/10.35111/wk4f-qt80} {\emph{English {G}igaword {F}ifth {E}dition}}.
\newblock Linguistic Data Consortium, Philadelphia, PA.
\newblock LDC2011T07.

\bibitem[{Pennington et~al.(2014)Pennington, Socher, and Manning}]{pennington-etal-2014-glove}
Jeffrey Pennington, Richard Socher, and Christopher Manning. 2014.
\newblock \href {https://doi.org/10.3115/v1/D14-1162} {{G}lo{V}e: Global vectors for word representation}.
\newblock In \emph{Proceedings of the 2014 Conference on Empirical Methods in Natural Language Processing ({EMNLP})}, pages 1532--1543, Doha, Qatar. Association for Computational Linguistics.

\bibitem[{Qi et~al.(2020)Qi, Zhang, Zhang, Bolton, and Manning}]{qi-etal-2020-stanza}
Peng Qi, Yuhao Zhang, Yuhui Zhang, Jason Bolton, and Christopher~D. Manning. 2020.
\newblock \href {https://doi.org/10.18653/v1/2020.acl-demos.14} {{S}tanza: A python natural language processing toolkit for many human languages}.
\newblock In \emph{Proceedings of the 58th Annual Meeting of the Association for Computational Linguistics: System Demonstrations}, pages 101--108, Online. Association for Computational Linguistics.

\bibitem[{Raffel et~al.(2020)Raffel, Shazeer, Roberts, Lee, Narang, Matena, Zhou, Li, and Liu}]{c4}
Colin Raffel, Noam Shazeer, Adam Roberts, Katherine Lee, Sharan Narang, Michael Matena, Yanqi Zhou, Wei Li, and Peter~J. Liu. 2020.
\newblock \href {http://jmlr.org/papers/v21/20-074.html} {Exploring the limits of transfer learning with a unified text-to-text transformer}.
\newblock \emph{Journal of Machine Learning Research}, 21(140):1--67.

\bibitem[{Shan et~al.(2023)Shan, Bauer, Carlson, and Manning}]{shan-etal-2023-english}
Alexander Shan, John Bauer, Riley Carlson, and Christopher Manning. 2023.
\newblock \href {https://doi.org/10.18653/v1/2023.findings-emnlp.788} {Do ``{E}nglish'' named entity recognizers work well on global {E}nglishes?}
\newblock In \emph{Findings of the Association for Computational Linguistics: EMNLP 2023}, pages 11778--11791, Singapore. Association for Computational Linguistics.

\bibitem[{Soldaini et~al.(2024)Soldaini, Kinney, Bhagia, Schwenk, Atkinson, Authur, Bogin, Chandu, Dumas, Elazar, Hofmann, Jha, Kumar, Lucy, Lyu, Lambert, Magnusson, Morrison, Muennighoff, Naik, Nam, Peters, Ravichander, Richardson, Shen, Strubell, Subramani, Tafjord, Walsh, Zettlemoyer, Smith, Hajishirzi, Beltagy, Groeneveld, Dodge, and Lo}]{dolma}
Luca Soldaini, Rodney Kinney, Akshita Bhagia, Dustin Schwenk, David Atkinson, Russell Authur, Ben Bogin, Khyathi Chandu, Jennifer Dumas, Yanai Elazar, Valentin Hofmann, Ananya~Harsh Jha, Sachin Kumar, Li~Lucy, Xinxi Lyu, Nathan Lambert, Ian Magnusson, Jacob Morrison, Niklas Muennighoff, Aakanksha Naik, Crystal Nam, Matthew~E. Peters, Abhilasha Ravichander, Kyle Richardson, Zejiang Shen, Emma Strubell, Nishant Subramani, Oyvind Tafjord, Pete Walsh, Luke Zettlemoyer, Noah~A. Smith, Hannaneh Hajishirzi, Iz~Beltagy, Dirk Groeneveld, Jesse Dodge, and Kyle Lo. 2024.
\newblock \href {https://arxiv.org/abs/2402.00159} {{Dolma: An Open Corpus of Three Trillion Tokens for Language Model Pretraining Research}}.
\newblock \emph{arXiv preprint arXiv:2402.00159}.

\bibitem[{Tjong Kim~Sang and De~Meulder(2003)}]{conll03}
Erik~F. Tjong Kim~Sang and Fien De~Meulder. 2003.
\newblock \href {https://www.aclweb.org/anthology/W03-0419} {Introduction to the {C}o{NLL}-2003 shared task: Language-independent named entity recognition}.
\newblock In \emph{Proceedings of the Seventh Conference on Natural Language Learning at {HLT}-{NAACL} 2003}, pages 142--147.

\bibitem[{Vallebueno et~al.(2024)Vallebueno, Handan-Nader, Manning, and Ho}]{vallebueno-etal-2024-statistical}
Andrea Vallebueno, Cassandra Handan-Nader, Christopher~D Manning, and Daniel~E. Ho. 2024.
\newblock \href {https://doi.org/10.18653/v1/2024.emnlp-main.510} {Statistical uncertainty in word embeddings: {G}lo{V}e-{V}}.
\newblock In \emph{Proceedings of the 2024 Conference on Empirical Methods in Natural Language Processing}, pages 9032--9047, Miami, Florida, USA. Association for Computational Linguistics.

\end{thebibliography}
\bibliographystyle{acl_natbib}

\end{document}